\documentclass[manuscript,screen]{acmart}
\usepackage{lineno}
\usepackage[utf8]{inputenc}
\usepackage{longtable}
\usepackage{subcaption}
\usepackage{lscape}
\usepackage{xcolor}
\usepackage{soul}

\usepackage{tabularx}
    \newcolumntype{L}{>{\raggedright\arraybackslash}X}

\definecolor{darkcerulean}{rgb}{0.03, 0.27, 0.49}

\usepackage{adjustbox}
\usepackage{array}

\newcolumntype{R}[2]{%
    >{\adjustbox{angle=#1,lap=\width-(#2)}\bgroup}%
    l%
    <{\egroup}%
}
\newcommand*\rot{\multicolumn{1}{R{45}{1em}}}

\AtBeginDocument{%
  \providecommand\BibTeX{{%
    \normalfont B\kern-0.5em{\scshape i\kern-0.25em b}\kern-0.8em\TeX}}}

\setcopyright{acmcopyright}
\copyrightyear{2019}
\acmYear{2019}
\acmDOI{}

\acmBooktitle{ACM Transactions on Human-Robot Interaction (THRI)}

\begin{document}

\title{Explainable Embodied Agents Through Social Cues: A Review}


\author{Sebastian Wallkötter}
\authornote{Both authors contributed equally to this research.}
\email{sebastian.wallkotter@it.uu.se}
\orcid{}
\author{Silvia Tulli}
\authornotemark[1]
\email{silvia.tulli@gaips.inesc-id.pt}
\affiliation{%
  \institution{\small Uppsala University and INESC-ID - Instituto Superior T\'ecnico}
  \streetaddress{}
  \city{}
  \state{}
  \postcode{}
}

\author{Ginevra Castellano}
\affiliation{%
  \institution{\small Uppsala University}
  \city{}
  \country{Sweden}
}

\author{Ana Paiva}
\affiliation{%
 \institution{\small INESC-ID and Instituto Superior T\'ecnico}
 \streetaddress{}
 \city{Lisbon}
 \state{}
 \country{Portugal}}

\author{Mohamed Chetouani}
\affiliation{%
  \institution{\small Institute for Intelligent Systems and Robotics, CNRS UMR 7222, Sorbonne Universit\'e}
  \streetaddress{}
  \city{Paris}
  \country{France}}
\email{}

\renewcommand{\shortauthors}{Wallkötter and Tulli, et al.}

\begin{abstract}
The issue of how to make embodied agents explainable has experienced a surge of interest over the last three years, and, there are many terms that refer to this concept, e.g., transparency or legibility. One reason for this high variance in terminology is the unique array of social cues that embodied agents can access in contrast to that accessed by non-embodied agents. Another reason is that different authors use these terms in different ways. Hence, we review the existing literature on explainability and organize it by (1) providing an overview of existing definitions, (2) showing how explainability is implemented and how it exploits different social cues, and (3) showing how the impact of explainability is measured.
Additionally, we present a list of open questions and challenges that highlight areas that require further investigation by the community.
This provides the interested reader with an overview of the current state-of-the-art.

\end{abstract}


\maketitle

\section{Introduction}
Embodied agents are capable of engaging in face-to-face interaction with humans through both, verbal and non-verbal behaviour. They are employed in situations in which joint activities occur, requiring teammates to be able to perceive, interpret, and reason about intentions, beliefs, desires, and goals to perform the right actions. However, even if an embodied agent is endowed with communicative behaviours, merely having them does not guarantee that its actions are understood correctly. 

Enabling an embodied agent to use its own behaviours to be better understood by human partners is studied as explainability (among other terms) and is an active research area (see figure \ref{fig:paper_histogram}). Embodied agents have access to differing social cues compared to their non-embodied counterparts, which sets them apart from artificial intelligence (AI) systems that are not embodied. These social cues are handy, because users - particularly non-experts - often have limited understanding of the underlying mechanisms by which an embodied agents choose their actions. Revealing such mechanisms builds trust and is considered ethical in AI \cite{euethics2019} and robotics alike. Embodiment-related social cues may help in this context to efficiently communicate such underlying mechanisms for a better interaction experience. Consequently, it is interesting to investigate explainability with a special focus on embodiment.

In human-robot interaction (HRI), embodiment often refers to the use of physical embodiment; however, testing a single aspect of the interaction in online studies is also common, such as the observation of the agent behavior in a simulation or the intepretation of its textual explanations. While this might introduce a confounding difference between the settings (e.g., not accounting for the effect of the embodiment) \cite{li2015}, in such studies, the risk of introducing a confounding factor is outweighed by the possibility isolating the aspect of interest. In explainability research studies with virtual agents manipulating a single behaviour happens frequently, and typically with the aim of adding the investigated aspect to a physically embodied agent later during the research. Hence, we chose to not only review physically embodied agents but also include virtual studies.

Although the concept of explainable embodied agents has become increasingly prevalent, the idea of explaining a system's action is not new. Starting in the late 1970s scholars already began to investigate how expert systems \cite{warren1977, scott1977, vanmelle1978} or semantic nets \cite{wick1992, georgeff1999}, which use classical search methods, encode their explanations in human readable form. Two prominent examples of such systems are MYCIN \cite{vanmelle1978}, a knowledge-based consultation program for the diagnosis of  infectious diseases, and PROLOG, a logic-based programming language. MYCIN is particularly interesting because it has already explored the idea of interactive design to allow both, inquiry about the decision in the form of specific questions, and rule-extraction based on previous decision criteria \cite{vanmelle1978}.

Connecting to this history, explainable embodied agents can be considered a subcategory of explainable AI systems, as they use techniques similar to those mentioned above but with the aim of interacting autonomously with humans and the environment. Therefore, explainable embodied agents should be able to explain their reasoning (justification), explain their view of the environment (internal state) or explain their plans (intent). These abilities can, for example, support the collaboration between agents and humans or improve the agent's learning by aiding the human teacher in selecting informative instruction \cite{chao2010}. Furthermore, explainable AI techniques might exploit attention direction to communicate points of confusion for embodied agents using gaze and verbalization \cite{perlmutter2016}. Some examples in the literature have proposed generating explanations of reasoning processes using natural language templates \citet{wang2016a}. In contrast, other work focuses on making the actions of the embodied agents explicit by design (i.e., legible) \cite{dragan2013b}.

In this paper, we want to show how these unique social cues can be used for building explainable embodied agents, and highlight which aspects require further inquiry. As such, \textbf{we contribute} to the field of explainability with \textbf{a review of the existing literature on explainability} that organizes it by (1) providing an overview of existing definitions, (2) showing how explainability is implemented and how it exploits different social cues, and by (3) showing how the effect of explainability is measured. This review aim to provide interested readers with an overview of the current state-of-the-art. Additionally, we present \textbf{a list of open questions and challenges} that highlight areas that require further investigation by the community. 

\subsection{Similar Reviews}
Other authors have previously written about explainability in the form of position papers and reviews \cite{spagnolli2018, theodorou2016, lyons2013, jacucci2014, fischer2018, felzmann2019}.

\citet{doshi2017} and \citet{lipton2018} sought to refine the discourse on interpretability by identifying the desiderata and methods of interpretability research. Their research focused on the interpretation of machine learning systems from a human perspective and identified trust, causality, transferability, informativeness, and fair and ethical decision-making as key aspects of their investigation. \citet{rosenfeld2019} provided a notation for defining explainability in relation to related terms such as interpretability, transparency, explicitness, and faithfulness. Their taxonomy encompasses the motivation behind the need for explainability (unhelpful, beneficial, critical), the importance of identifying the target of the explanation (regular user, expert user, external entity), when to provide the explanation, how to measure the effectiveness of the explanation and which interpretation of the algorithm has been used. 
\citet{anjomshoae2019} conducted a systematic review on the topic of explainable robots and agents and clustered the literature with respect to the user's demographics, the application scenario, the intended purpose of an explanation, and whether the study  was grounded in social science or had a psychological background. The review summarized the methods used to implement the explanation to the user with its dynamics (context-aware, user-aware, both or none), and the types of explanation modality. \citet{alonso2018} proposed a review of the system's explainability in a shared autonomy framework, stressing the role of explainability in flexible and efficient human-robot collaborations. Their review underlines how explainability should vary in relation to the level of system autonomy and how the exploitation of explainability mechanisms can result from an explanation, a property of an interface, or a mechanical feature.

Other reviews explored the cognitive aspects of explainability by targeting the understanding of behaviour explanations and how this helps people find meaning in social interaction with artificial agents. The book by \citet{malle2004} argued that people expect explanations using the same conceptual framework used to explain human behaviours. Similarly, the research of \citet{miller2019} focused on the definition of explanations in other relevant fields, such as philosophy, cognitive psychology/science, and social psychology. The author described an explanation as contrastive, selected and social, specifying that the most likely explanation is not always the best explanation for a person.

\subsection{Methodology}
For this review, we chose to use a keyword based search in Scopus\footnote{https://www.scopus.com/} database to identify relevant literature, as this method makes our search reproducible. 

First, we identified a set of relevant papers in an unstructured manner based on previous knowledge of the area. From each paper, we extracted both, the indexed and the author keywords, and rank ordered each keyword by occurrence. Using this method, we identified key search terms such as \textit{human-robot interaction, transparent, interpretable, explainable,} or \textit{planning}.

\begin{flushleft}
\small
\begin{table}[th]
\caption{Inclusion Criteria and Search String}
\label{tab:inclusion_criteria}
\begin{tabular}{@{}lp{.35\linewidth}p{.35\linewidth}@{}}
\toprule
Topic & Description & Search Term \\ \midrule
Human Involvement & Exclude papers without human involvement, e.g., position papers or agent-agent interaction & ("human-robot" OR "child-robot" OR "human-machine")\\ 
Explainability   & & (transparen* OR interpretabl* OR explainabl*)\\
Explainability II  & & (obser* OR legib* OR visualiz* OR (commun* AND "non-verbal"))\\
Autonomy     & Exclude papers that are not using an autonomous agent & (learn* OR plan* OR reason* OR navigat* OR adapt* OR personalis* OR decision-making OR autonomous) \\
Social Cues    & Exclude papers that do not have a social interaction between human and agent & (social OR interact* OR collab* OR shared OR teamwork OR (model* AND (mental OR mutual)))\\
Agent   & Exclude papers that do not use an agent & (agent* OR robot* OR machine* OR system*) \\
Recency & Only consider the last 10 years & ( LIMIT-TO (PUBYEAR, 2019) OR ... OR LIMIT-TO (PUBYEAR, 2009) ) \\
Subject Area   & Only consider papers from computer science, engineering, math, psychology, social sciences, or neuroscience & (LIMIT-TO (SUBJAREA, "COMP") OR LIMIT-TO (SUBJAREA, "ENGI") OR LIMIT-TO (SUBJAREA, "MATH") OR LIMIT-TO (SUBJAREA, "SOCI") OR LIMIT-TO (SUBJAREA, "PSYC") OR LIMIT-TO (SUBJAREA, "NEUR")) \\\\
Full Search String & \multicolumn{2}{l}{\begin{minipage}{0.7\linewidth}TITLE-ABS-KEY ("human-robot" OR "child-robot" OR "human-machine") AND (transparen* OR interpretabl* OR explainabl*) AND (obser* OR legib* OR visualiz* OR (commun* AND "non-verbal")) AND (learn* OR plan* OR reason* OR navigat* OR adapt* OR personalis* OR decision-making OR autonomous) AND (social OR interact* OR collab* OR shared OR teamwork OR (model* AND (mental OR mutual))) AND (agent* OR robot* OR machine* OR system*) AND (LIMIT-TO (PUBYEAR, 2019) OR LIMIT-TO (PUBYEAR, 2018) OR LIMIT-TO (PUBYEAR, 2017) OR LIMIT-TO (PUBYEAR, 2016) OR LIMIT-TO (PUBYEAR, 2015) OR LIMIT-TO (PUBYEAR, 2014) OR LIMIT-TO (PUBYEAR, 2013) OR LIMIT-TO (PUBYEAR, 2012) OR LIMIT-TO (PUBYEAR, 2011) OR LIMIT-TO (PUBYEAR, 2010) OR LIMIT-TO (PUBYEAR, 2009)) AND (LIMIT-TO (SUBJAREA, "COMP") OR LIMIT-TO (SUBJAREA, "ENGI") OR LIMIT-TO (SUBJAREA, "MATH") OR LIMIT-TO (SUBJAREA, "SOCI") OR LIMIT-TO (SUBJAREA, "PSYC") OR LIMIT-TO (SUBJAREA, "NEUR"))\end{minipage}} \\
\bottomrule
\end{tabular}
\end{table}
\end{flushleft}
\normalsize

We then grouped these keywords by topic (see table \ref{tab:inclusion_criteria}) and performed a pilot search on each topic to determine how many of the initially identified papers were recovered. We then combined each group using \textit{AND}, which led to a corpus of $263$ papers\footnote{A spreadsheet detailing the raw search results can be found in the supplementary files.}. All authors participated in this initial extraction process.

Next, we manually filtered this list to further remove unrelated work by judging relevance based on titles, abstracts, and full text reads. To ensure selection reliability, both main authors rated inclusion of each paper independently. If both labelled the paper as relevant, we included the paper; similarly, if both labelled it as unrelated, we excluded it. For papers with differing decisions, we discussed their relevance and made a joint decision regarding the paper's inclusion. This left us with $32$ papers for the final review \footnote{A spreadsheet detailing the reason for excluding a paper can be found in the supplementary files.}.

For the excluded papers, each main author indicated why a paper was excluded for the following reasons:
\begin{itemize}
    \item \textbf{no article} The paper was a book chapter or review paper. ($\sim 15.63\%$ excluded)
    \item \textbf{wrong topic} The paper presented work in a different focus area, e.g, material science, teleoperation, or generically making robots more expressive (without considering explainability). ($\sim 45.98\%$ excluded)
    \item \textbf{wrong language} The paper was not written in English. ($\sim 0.46\%$ excluded)
    \item \textbf{no embodiment} The paper did not consider an embodied agent. ($\sim 11.26\%$ excluded)
    \item \textbf{no autonomy} The paper did not consider autonomous embodied agents. ($\sim 13.33\%$ excluded)
    \item \textbf{no social interaction} The paper did not investigate explainability in a context where a human was present. ($\sim 13.10\%$ excluded)
\end{itemize}

\begin{figure}[t]
 \centering
 \includegraphics[width=.8\textwidth]{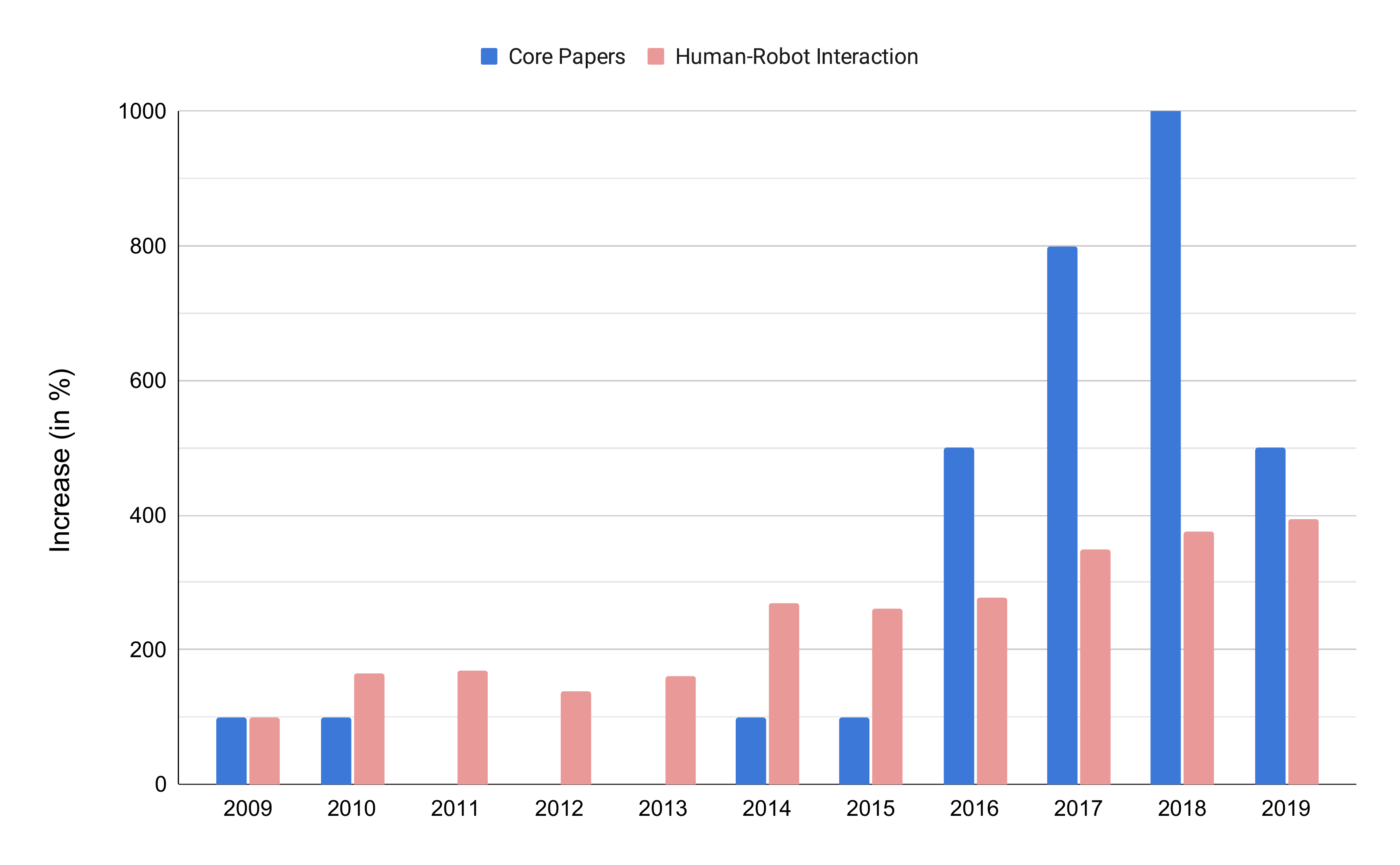}
 \caption{Percentage increase in the number of publications relative to the first year (2009) in the reviewed area. For comparison, the graph also shows the percentage increase in the number of publications overall measured by the number of publications indexed in Scopus that use the keyword "Human-Robot Interaction".}
 \label{fig:paper_histogram}
\end{figure}

Figure \ref{fig:paper_histogram} shows a comparison between the publication rate of the identified papers (core papers) and the publication rate of the general field, which is measured, as a proxy, by papers published using the keyword \textit{human-robot interaction}. We can see that there is a growing interest in the topic reviewed here.

\section{Definitions}\label{sec:definition}
\subsection{Review of Definitions}
Before starting to look into how different social cues are exploited, it is important to understand what different scholars mean when they talk about explainability.


Among the surveyed papers, we analyzed the terms used in the title. We extrapolated the root terms and categorized them (e.g. explanation -> explainability, expressive -> expressivity). Assuming that papers with the same authors share the same definition, we grouped them by author, and only report the most recent definition in Table \ref{tab:definition}. When the definition was from an outside source not cited before, we mentioned both the definition and the reference.

Some authors that refer to transparency identify transparency as the process or the capability of revealing hidden or unclear information about the robot. \citet{akash2018a}\cite{akash2018b} and \citet{ososky2014} cited the definition of \citet{chen2017}, which named transparency the descriptive quality of an interface. The descriptiveness of the interface affects the operator on three levels (perception, comprehension, and projection) leveraged by Endsley's model of situation awareness \cite{endsley1995}. \citet{ososky2014} further refer to the dictionary definition of \textit{transparency}\footnote{https://www.merriam-webster.com}; thus, they chose the property of being \textit{able to be seen through} or \textit{easy to notice or understand} as their definition. \citet{chao2010} gave a similar definition in the context of robot active learning referring to transparency as revealing to the teacher unknown information about the robot.


In an extension, \citet{floyd2016} and \citet{hayes2017} refer to both explainability and transparency. \citet{hayes2017} describe explainability as the embodied agent's ability to synthesize policy descriptions and respond to human collaborators.

Starting from the research of \citet{kulkarni2016} and \citet{zhang2009}, \citet{chakraborti2017} and \cite{sreedharan2017} formulated their idea of explainability as a model reconciliation problem. They use explanations to move the human model of the robot to be in conformance with the embodied agent model. \citet{gong2018} differentiate their work by shifting the interest in signalling robot intentions before actions occur. Along similar lines, \citet{tabrez2019} defined explainability as a policy update given by the robot to the human to reduce the likelihood of costly or dangerous failures during joint task execution.

\citet{baraka2018a} employed the term \textit{expression} for \textit{externalizing hidden information of an agent}. The work of \citet{kwon2018} extended this notion of expressivity by targeting the communication of robot incapability. To do so, the robot should reveal the intentions and the cause of its failures. The same concept is referred to using the word communicability; e.g., \citet{huang2012} refer to \textit{communicate} for describing the robot capability of expressing its objectives and the robots capability to enable end-users to correctly anticipate its behaviour.

Similarly, \citet{schaefer2016} investigated the \textit{understandability} of the embodied agent's intentions for effective collaborations with humans. Following this idea, \citet{grigore2018} referred to \textit{predictability}, building upon hidden state representation.

Other studies in the literature do not refer to a specific capability of their system in the title but highlighted the application scenarios (e.g., autonomous driving \cite{brown2017}, human robot teams in the army \cite{evans2017}, interactive robot learning \cite{lutkebohle2009}) or mentioned \textit{behavioural dynamics} \cite{lamb2018} and \textit{human-machine confidence} \cite{legg2019}.

Although there exists large diversity and inconsistency in the language, there seem to be commonalities in what the authors identify as explainability. Furthermore, other authors that use terms such as transparency, expressivity, understandability, predictability and communicability present definitions that are congruent with those provided for explainability. We noticed that all the given definitions share the following aspects: (1) they all refer to an embodied agent's capability or system's module, (2) they all specify that what should be explained/signalled are the internal workings of the robot (e.g. agent's intent, policy, plan, future plans), and (3) they all consider the human as a target of the explanations. 

\subsection{Our Definition}
Therefore, we provide a definition that aims to be comprehensive for our literature. \textbf{We define the explainability of embodied social agents as their ability to provide information about their inner workings using social cues, such that an observer (target user) can infer how/why the embodied agent behaves the way it does.}


\begin{table}[t]
\centering
\caption{Papers by Terminology Used in the Title}
\label{tab:terminology}
\begin{tabular}{@{}lp{.5\linewidth}@{}}
\toprule
Category & Paper \\ \midrule
Transparency &  \cite{akash2018a,akash2018b,boyce2015,chao2010,chen2017,fischer2018,floyd2016,perlmutter2016,poulsen2018,roncone2017} \cite{floyd2017, hayes2017}\\\\
Explainability &  \cite{chakraborti2017,gong2018,sreedharan2017,tabrez2019,wang2016a,wang2016b, wang2018}, \cite{floyd2017, hayes2017}\\\\
Expressivity & \cite{zhou2017,kwon2018,baraka2018a} \\\\
Understandability & \cite{schaefer2016,sciutti2014} \\\\
Predictability & \cite{sheikholeslami2018,grigore2018} \\\\
Communicability & \cite{huang2019,lee2019} \\\\
None & \cite{brown2017,lamb2018,legg2019,lutkebohle2009,khoramshai2019}\\ \bottomrule
\end{tabular}
\end{table}


\subsection{Motivation}
While investigating the definitions, we noticed that the motivation of the experiment plays a key role in the choice of a specific definition. In particular, we identified the following reasons for investigating explainability:

\begin{itemize}
    \item \textbf{Interactive machine/robot learning} investigates the need of explainability in the context of robot learning. The main idea is that revealing the embodied agent's internal states allows the human teacher to provide more informative examples \cite{lutkebohle2009, chao2010, thomaz2008, tabrez2019}. 
    \item \textbf{Human Trust} states that adding explainability increases human trust and system reliability. This motivation empathizes the importance of communicating the agent's uncertainty, incapability, and existence of internally conflicting goals \cite{wang2016b, roncone2017, kwon2018, schaefer2016}. 
    \item \textbf{Teamwork} underlines the value of explainability in human-robot collaboration scenarios to build shared mental models and predict the embodied agent's behaviour \cite{huang2019, legg2019, chakraborti2017, hayes2017, sciutti2014}. 
    \item \textbf{Ethical decision-making} suggests that communicating the embodied agent's decision-making processes and capabilities, paired with situational awareness, increases a user's ability to make good decisions \cite{poulsen2018, kwon2018, akash2018a}. 
\end{itemize}

We have aggregated the individual definitions used in each paper and the motivations behind them in table \ref{tab:definition}. Guidance and dialogue with a human tutor are aspects that are important for interactive machine/robot learning ("dialog to guide human actions" \cite{lutkebohle2009}, "revealing to the teacher what is known and what is unclear" \cite{chao2010}). Providing information about the level of uncertainty and expressing robot incapability are core concepts of explainability that enhance human trust ("communicate uncertainty" \cite{wang2016b}, "provide appropriate measures of uncertainty" \cite{roncone2017}, "express robot incapability" \cite{kwon2018}).
The ability to anticipate an embodied agent's behaviour and establish a two-way collaborative dialogue by identifying relevant differences between the humans' and the robots' model are shared elements of the definitions around teamwork ("anticipate robot's behaviour" \cite{huang2019}, "establish two-way collaborative dialogue" \cite{legg2019}, "reconcile the relevant differences between the humans' and robot's model" \cite{chakraborti2018b}, "share expectations" \cite{hayes2017}). Authors that refer to ethical decision-making identify the communication of intentions and context-dependent recommendations as crucial information ("robot's real capabilities, intentions, goals and limitations" \cite{poulsen2018}, "context-dependent recommendations based on the level of trust and workload" \cite{akash2018b}).

\section{Social Cues}
\label{sec:communication_modality}

\begin{table}[t]
\centering
\caption{Papers on Explainability Ordered by Social Cues}
\label{tab:communication_modalities}
\begin{tabular}{@{}lp{.5\linewidth}@{}}
\toprule
Category & Paper \\ \midrule
Speech & \cite{fischer2018, lutkebohle2009, perlmutter2016, roncone2017, tabrez2019}\\
Text & \cite{akash2018a, akash2018b, chen2017, floyd2016, floyd2017, gong2018, hayes2017, poulsen2018, sreedharan2017, wang2016a, wang2016b, wang2018}\\
Movement & \cite{brown2017, chao2010, huang2019, kwon2018, lamb2018, lutkebohle2009, perlmutter2016, sciutti2014, sreedharan2017, zhou2017}\\
Imagery & \cite{boyce2015, brown2017, legg2019, perlmutter2016, roncone2017, schaefer2016}\\
Other/Unspecified & \cite{baraka2018a, perlmutter2016}/ \cite{chakraborti2017, khoramshai2019, lee2019, grigore2018}* \\
\bottomrule
\end{tabular}
\end{table}

We have claimed above that embodied agents can become explainable using unique types of social cues that are not available to agents lacking such embodiment. One example is the ability to point to important objects in a scene - assuming the agent has an extremity that affords pointing. A non-embodied agent has to use a different way to communicate the importance of that object.

Hence, we screened the core papers to check which modality the authors deployed to make the agent more explainable. We then logically grouped the core papers based on these types of social cues and identified five groups:

\begin{itemize}
\item \textbf{Speech} A lexical statement uttered verbally using a text-to-speech mechanism.
\item \textbf{Text} A lexical statement displayed as a string presented as an element of a typically screen-based user interface.
\item \textbf{Movement} A movement that is either purely communicative, or that alters an existing movement to make it more communicative.
\item \textbf{Imagery} A drawing or image (often annotated) presented as an element of a user interface (typically screen-based).
\item \textbf{Other/Unspecified} All papers that use  social cues that do not fit within above set of categories, or where the authors did not explicitly specify the modality (the latter is marked with an asterisk*).
\end{itemize}

This grouping is shown in table \ref{tab:communication_modalities}. Surprisingly, our search did not yield any papers that investigate non-lexical utterances (beeping noise, prosody, etc.), which was contrary to our expectations. A possible explanation for this could be that our search terms did not capture a broad enough scope, because experiments investigating such utterances may use yet again a different terminology. Another possibility could be that it seems much harder to communicate an explanation through \textit{beeps and boops} instead of using speech; especially when considering the wide availability of text-to-speech synthesizers (TSSs).

The wide availability of TTS synthesizers may also explain another interesting result of this analysis. Many works focus on lexical utterances (speech and text). Potentially, such utterances are seen as easier to work with when giving explanations because of the high expressivity of natural language.

On the other hand, lexical utterances may add additional complexity to the interaction, because a sentence has to be interpreted and understood, whereas, other social cues may be faster/easier to interpret. While there does exist work that investigates the added cognitive load of having explainability versus not having explainability \cite{wang2016b}, comparing lexical utterances with other forms of explainability is currently underexplored. This prompts the question of whether lexical utterances are always superior to achieve explainability and, if not, under what circumstances other social cues perform better.

\section{Explainability Mechanisms} \label{sec:implementation_methods}
Next, we report and discuss the methods employed to achieve explainable behaviours in social agents with a specific focus on embodied agents. From an HRI perspective, introducing explainability mechanisms is challenging, as uncertainty is inherent to the whole process of interaction from perception to decision and action. In addition, the methods used to implement explainability require explicit consideration of the human capability to correctly infer the agent goals, intentions or actions from the observable cues.  

Looking at one specific implementation, \citet{thomaz2006a} introduced the socially guided machine learning (SG-ML) framework which seeks to augment traditional machine learning models by enabling them to interact with humans. Two interrelated questions are investigated: (1) how do people want to teach embodied agents and (2) how do people design embodied agents that learn effectively from natural interaction and instruction. This framework considers a reciprocal and tightly coupled interaction; the machine learner and human instructor cooperate to simplify the task for each other. SG-ML considers explainability to be a communicative act that helps humans understand the machine's internal state, intent, or objective during the learning process. 

\begin{figure}
    \centering
    \frame{\includegraphics[scale=0.47]{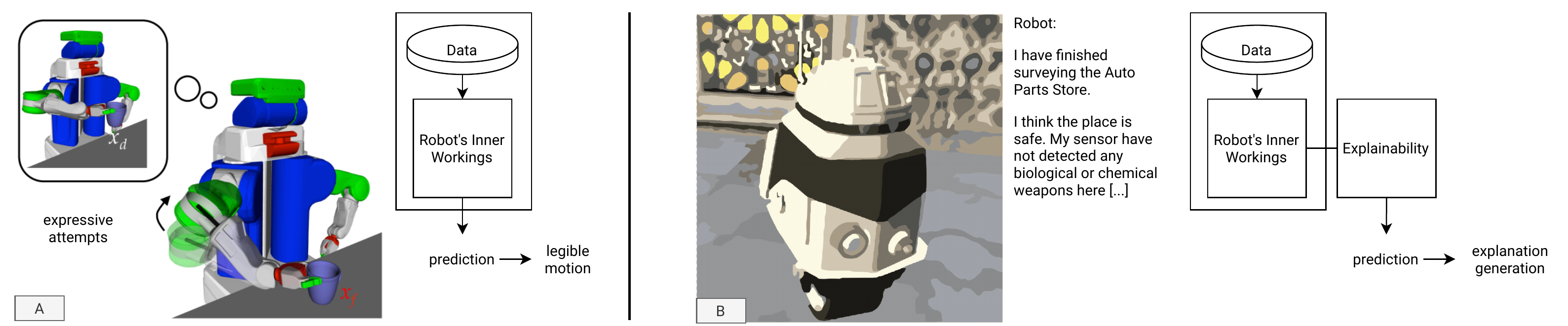}}
    \caption{Example A - \citet{kwon2018}, Example B - inspired by \citet{wang2016a}. Agent and Explainability Mechanisms. Intrinsic explainability refers to models that are explainable by design. Post-hoc (external) explainability refers to explainability mechanisms that are applied after the decision or execution of actions. Explainability could be either performed by external mechanisms that are separable from the task execution (visualization) or intrinsically computed by the agent policy (e.g. query learning, communicative gestures). Permission to reuse figure A was kindly provided by the authors.} 
    \label{fig:ML-transparency}
\end{figure}

\paragraph{Interactive situations}
As humans interact with robots, explainability becomes a key element. Different interactive scenarios have been explored in the analysed papers, in particular, scenarios where humans shape the behaviour of a robot by providing instructions and/or social cues through interactive learning techniques. Despite of a strong emphasis on interactive robot learning scenarios, there are also other types of tasks as illustrated  by Figure \ref{fig:ML-transparency}. For example, the behaviour shaping explored in  \cite{knox2009, najar2019} aims to exploit instructions and/or social cues to steer robot actions towards desired behaviours. Various interaction schemes have been proposed including instructions \cite{grizou2013, paleologue2017}, advice \cite{griffith2013}, demonstrations \cite{argall2009}, guidance \cite{suay2011, najar2016}, and evaluative feedback \cite{knox2013,najar2016}. Then, computational models, mostly based on machine learning, are exploited to modify agent states $s$ and actions $a$ to achieve a certain goal $g$. 

As mentioned by \citet{broekens2019}, most of computational approaches for social agents consider a primary task, e.g., learning to pick an object, and explainability arises as a secondary task by either communicating the agent's internal states, intentions, or future goals. Existing works distinguish the nature of actions performed by the agent, such as task - oriented actions $a_{\mathbf{T}}$ and communication - oriented actions $a_{\mathbf{C}}$. $a_{\mathbf{T}}$ are used to achieve a goal $g$ such as sorting objects. $a_{\mathbf{C}}$ are used by the agent to communicate with humans such as queries or pointing to objects. This follows from the speech act theory \cite{koller1970}, which treats communication as actions that have an intent and an effect (a change of mind by the receiver of the communication).

In such a context, explainability mechanisms are employed to reduce uncertainty during the shaping process using communicative actions $a_{\mathbf{C}}$ before, during, or after performing a task action $a_{\mathbf{T}}$, which will change the agent's next state $s'$. The challenge for explainability mechanisms is then to transform agent states $s$ and task oriented actions $a_{\mathbf{T}}$ into communicative actions using either using natural language or non-verbal cues. To tackle this challenge, several explainability mechanisms have been proposed for embodied agents.

\paragraph{Computational paradigms}
Various computational paradigms are employed ranging from supervised learning to reinforcement learning.
In supervised learning, a machine is trained using data, e.g., different kinds of objects, which are labelled by a human supervisor. In the case of interactive robot learning, the supervisor is a human teacher and provides labelled examples based on embodied agent queries or explanations. In addition, the level of expertise of the human is rarely questioned and considered ground truth. To tackle these challenges, \citet{chao2010} proposed an active learning framework for teaching embodied agents to classify pairs of objects ($a_{\mathbf{T}}$). Active learning is a type of interactive machine learning that allows the learner to interactively query the supervisor to obtain labels for new data ($a_{\mathbf{C}}$). By doing so, the robot improves both learning and the explainability by communicating about uncertainty. Often, active learning is a form of semi-supervised learning, that combines human supervision and processing of unlabelled data.   

Another paradigm is reinforcement learning (RL) \cite{sutton2018reinforcement}, which is one of the three basic machine learning paradigms. Here, an agent acts in an environment, observing its state $s$ and receiving a reward $r$. Learning is performed by a trial-and-error process through interaction with the environment and leverages the Markov decision process (MDP) framework. MDPs are used to model the agent's policy and help with decision making under uncertainty. This paradigm allows one to represent, plan, or learn an optimal policy - a mapping from current state to action. Analysing this policy provides insights into future and current states and actions. \citet{hayes2017} developed a policy explanation framework based on the analysis of execution traces of an RL agent. The method generates explanations ($a_{\mathbf{C}}$) about the learned policy ($a_{\mathbf{T}}$) in a way that is understandable to humans. In RL, theoretical links between (task) learning schemes and emotional theories could be performed. \citet{broekens2019} investigated how temporal difference learning \cite{sutton2018reinforcement} could be employed to develop an emotionally expressive learning robot that is capable of generating explainable behaviours via emotions. 

To increase the understanding of robot intentions by humans, the notion of legibility is often introduced in robotics. Legibility and explainability are considered similar notions that aim to reduce ambiguity over possible goals that might be achieved. One key concept for achieving legibility/explainability is to explicitly consider a model of the human observer, and find plans that disambiguate possible goals. \citet{dragan2013b} proposed a mathematical model able to distinguish between legibility and predictability. Legibility is defined as the ability to anticipate the goal, and predictability, is defined as the ability to predict the trajectory. The mathematical model exploits observer expectations to generate legible/explainable plans. \citet{huang2019} propose modelling how people infer objectives from observed behaviour, and then selecting those behaviours that are maximally informative. Inverse reinforcement learning is used to model observer capability of inferring intentions from the observation of agent behaviours. Explainability implementations based on these methods consider that task-oriented actions ($a_{\mathbf{T}}$) and communicative actions ($a_{\mathbf{C}}$) are performed through the same channel, e.g., movement of the robot's arm both achieves a task and communicates the goal \cite{sheikholeslami2018}, \cite{sciutti2014}.  

\paragraph{Explainability mechanisms}
In artificial intelligence (AI), explainability addresses the understanding of the mechanisms by which a model works aiming to reduce the model's \textit{black box} \cite{arrieta2019}. Deep learning is a typical black-box machine learning method that achieves data representation learning using multiple nonlinear transformations. In contrast, a linear model is considered as explainable since the model is fully understandable and explorable by means of mathematical analysis and methods. In \citet{arrieta2019}, the authors argue that a model is considered to be explainable if by itself it is understandable and propose various levels of model explainability: (1) simulatability, the ability to be simulated or conceptualized strictly by a human, (2) decomposability, the ability to explain each part of the model, and (3) algorithmic transparency, the ability of the user to understand the process followed by the model to produce any given output from its input data.    

Intrinsic explainability refers to models that are explainable by design (Figure \ref{fig:ML-transparency}). Post-hoc (external) explainability refers to explainability mechanisms that are applied after the decision or execution of actions. Post-hoc methods are decoupled from the model and aim to enhance the explainability of models that are not explainable by design (intrinsic) \cite{lipton2018}. Post hoc explainability mechanisms such as visualization, mapping the policy to natural language, or explanation are used to convert a non-explainable model into a more explainable one. 

A large body of work aiming to achieve explainability in human-agent interaction does not explicitly refer to definitions that originate from machine learning. Explainability can be either performed by external mechanisms that are separable from the task execution (visualization) \cite{sciutti2014,perlmutter2016,zhou2017} or intrinsically computed by the agent policy (e.g. query learning, communicative gestures) \cite{chao2010,sheikholeslami2018}.

The implementation of explainability can also be done at several levels. For example, via the situation-awareness-based agent transparency (SAT) model, which is based on a Belief, Desire, Intention (BDI) architecture, considers three levels of explainability: Level 1–Basic Information (current status/plan); Level 2–Reasoning Information; Level 3–Outcome Projections \cite{boyce2015}.

Mapping agent policy ($a_{\mathbf{T}}$) to natural language ($a_{\mathbf{C}}$) is a methodology that is increasingly employed in AI to design explainable AI \cite{arrieta2019}. In HRI, a similar trend is observed \cite{akash2018b,hayes2017,tabrez2019,wang2016b}. The challenge will be to map agent policy to both verbal and non-verbal cues (see also section \ref{sec:communication_modality}).

\section{Evaluation Methods}\label{sec:evaluation_methods}
Existing works assess the effects of explainability on a variety of measures including but not limited to, self-reported understanding of the agent \cite{gong2018}, number of successful task completions \cite{wang2016b}, number of false decisions \cite{wang2016b}, task completion time \cite{chao2010}, number of irredeemable mistakes \cite{tabrez2019} or trust in automation \cite{boyce2015}. During our review three major categories of measurements emerged:

\begin{itemize}
    \item \textbf{Trust} measures how willing a user is to agree with a decision of a robot - based on the provided explanation -, how confident a user is about the embodied agent's internal workings (internal state), or if the user agrees with the plan provided by the robot (intent). It is measured using a self-report scale.
    \item \textbf{Robustness} measures the avoidance of failure during the interaction. Typically researchers want to determine whether the embodied agent's intent has been communicated correctly. It is often measured observationally, e.g., by counting the frequency of successful achievements of a goal.
    \item \textbf{Efficiency} measures how quickly the task is completed. The common hypothesis behind using this measure is that the user can adapt better to a more explainable robot, and form a more efficient team. It is commonly measured by wall clock time, or number of steps until the goal.
\end{itemize}

Among these measures, trust received the most attention. While there is large variance in which scale is used (often scales are self-made), a common element in the studies is the use of self-report questionnaires.

Although the consensus is that the presence of explainability generally increases trust (see Table \ref{tab:measures-overview}), how effective a particular social cue is in doing so has received much less attention. Comparisons that do exist often fail to find a significant difference between them \cite{wang2016b, boyce2015}. Similarly, due to the large range of mechanisms tested - and the even larger array of scenarios -, there is little work on how robust a specific mechanism performs across multiple scenarios. Hence, while some form of explainability seems to be clearly better then none, which specific mechanism to choose for which specific situation remains an open question.

Less studied, but no less important, is the effect of explainability on the robustness of an interaction. Research on the interplay between explainability and robustness uses tasks where mistakes are possible, and measures how often these mistakes occur \cite{boyce2015, perlmutter2016}. The core idea is that participants create better mental models of the robot when it is using explainability mechanisms. Better models will lead to better predictions of the embodied agent's future behaviour, allowing participants to anticipate low performance of the robot, and to avoid mistakes in task execution. However, experimental evidence on this hypothesis is not always congruent, with the majority of studies showing support for the idea, e.g., \cite{perlmutter2016}, and other studies finding no significant difference, e.g., \cite{boyce2015}. As the majority does find a positive effect, we can conclude that explainability does help improve reliability, although not in all circumstances. A more detailed account of when it does or does not remains a subject for future experimental work.

Finally, efficiency is a metric that some researchers have considered while manipulating explainability. It has been operationalized by comparing wall clock time until task completion across conditions \cite{chao2010}, or time until human response \cite{chen2017}. Of the three types of measures, this type has received the least attention, and the findings are quite mixed. Approximately half of the analysed papers find that making embodied agents explainable makes the team more efficient, while the other half find no difference. However, a clear explanation for these conflicting findings remains a topic of future work.

\begin{table}[t]
\centering
\caption{Papers on Explainability by Measure}
\label{tab:measures-overview}
\begin{tabular}{@{}lll@{}}
\toprule
Type & Outcome & Papers \\ \midrule
Robustness & Positive & \cite{chen2017, wang2018, wang2016b, kwon2018, sciutti2014, lamb2018, huang2019, perlmutter2016, baraka2018b, grigore2018} \\
Robustness & Negative &  \\
Robustness & Non-significant & \cite{chao2010} \\
Robustness & No statistical test & \cite{legg2019, sreedharan2017, sheikholeslami2018} \\
Trust & Positive & \cite{chao2010, chen2017, lee2010, boyce2015, baraka2018b, zhou2017, wang2016a, wang2018, schaefer2016} \\
Trust & Negative &  \\
Trust & Non-significant & \cite{akash2018b} \\
Trust & No statistical test & \cite{akash2018a, chakraborti2017, kwon2018, tabrez2019, poulsen2018, floyd2016, floyd2017, gong2018} \\
Efficiency & Positive & \cite{lee2010, wang2016b, akash2018b} \\
Efficiency & Negative &  \\
Efficiency & Non-significant & \cite{perlmutter2016, wang2016a, chen2017, chao2010} \\
Efficiency & No statistical test & \cite{roncone2017, chakraborti2017} \\
other & any & \cite{lutkebohle2009, evans2017, chakraborti2018a, khoramshai2019, hayes2017} \\ \bottomrule
\end{tabular}
\end{table}

Table \ref{tab:measures-overview} shows the core papers grouped by the evaluation methods discussed above and indicates whether the effect of explainability on it was positive, negative, or non-significant. One important note is that many papers introduce a measurement called \textit{accuracy}; however, usage of this term differs between authors. For example, \citet{chao2010} used accuracy to refer to the embodied agent's performance after a teaching interaction; hence it was being a measure of robustness, whereas \citet{baraka2018a}'s accuracy referred to people's self-rated ability to predict the robot's move correctly, a measure of trust.

In summary, there is enough evidence that explainability offers a clear benefit to virtual embodied agents in building trust, with some support for physical embodied agents. Additionally, there is evidence that explainability can decrease the chance of an unsuccessful interaction (improve robustness). However, papers looking to improve the efficiency of the interaction find mixed results. A possible explanation for this could be that while explainability makes the interaction more robust, the time added for the embodied agents to display and for the human to digest the additional information nullifies the gain in efficiency. 

In addition to the above analysis, this section identified the following open questions: (1) Is a particular explainability mechanism best suited for a specific type of embodied agent, a specific type of scenario, or both? (2) What are good objective measures with which we can measure trust in the context of explainability? (3) Why does explainability have a mixed impact on the efficiency of the interaction?

\section{Discussion}
In the above sections we provided a focused view on four key aspects of the field: (1) definitions used, and the large diversity thereof, (2) which social cues and (3) algorithms are used to link explainability mechanisms to the embodied agent's state or intent, and (4) the measurements to assess explainability mechanisms. What is missing is a discussion of how these aspects relate to each other when looked at from a $10,000$ foot view, and a discussion of the limitations of our work.

It is almost self-explanatory that the scenario chosen to study a certain explainability mechanism depends on the author's research goal. As such, it is unsurprising that we can find a large diversity of tasks, starting from evaluation in pure simulation \cite{hayes2017}, or discussions of hypothetical scenarios \cite{lee2010,poulsen2018} all the way to joint furniture assembly \cite{roncone2017}.

The most dominant strand of research has its origin in decision making, and mainly views the robot as a support for human decisions \cite{akash2018a,  akash2018b, chen2017, floyd2016, floyd2017, boyce2015, wang2018, wang2016b, wang2016a}. In this line of research, explainability is mostly commonly defined via the SAT-model (i.e., Situation Awareness-Based Agent) \cite{chen2018}. One of the key questions is how much a person will trust the embodied agent's suggestions, based on how detailed the given justification for the embodied agent's decision is. While these studies generally test a virtual agent shaped like a robot, the findings here can be easily generalized to the field of human-computer-interaction (HCI), due to their design. Hence, SAT model-based explanations can help foster trust not only in HRI, but also in the domain of expert systems and AI. Hence, this work partially overlaps with the domain of explainable AI (XAI).

The second strand of research sets itself apart by using humans as pure observers \cite{chakraborti2017, gong2018,huang2019, sreedharan2017, sciutti2014, baraka2018b, zhou2017, sheikholeslami2018, kwon2018, chakraborti2018a}. Common scenarios focus on communicating the embodied agent's internal state or intent by having humans observe a physical robot \cite{sheikholeslami2018, sciutti2014, baraka2018b} or video recordings/simulations of them \cite{zhou2017, baraka2018b, kwon2018}. Other researchers choose to show maps of plans generated by the robot and explanations thereof \cite{chakraborti2017, gong2018, huang2019, sreedharan2017}; the researchers' aim here is to communicate the robot's intent. In all scenarios, the goal is typically to improve robustness, although other measures have been tested. 

Particularly well done here is the work of \citet{baraka2018b}, who first describe how to enhance a robot with LED lights to display its internal state, use crowd sourcing to generate expressive patterns for the LEDs, and then validate the pattern's utility in both a virtual and a physical user study. This pattern of having participants - typically from Amazon Mechanical Turk (AMT) - generate expressive patterns in a first survey, and then validate them in a follow-up study was also employed by \citet{sheikholeslami2018} in a pick-and-place scenario. We think that this crowdsourcing approach deserves special attention, as it will likely lead to a larger diversity of candidate patterns compared to an individual researcher generating them. Considering the wide availability of online platforms, such as AMT and Polific, this is a tool that future researchers should leverage.

A third strand of research investigates explainability in interaction between a human and a robot \cite{schaefer2016, lamb2018, roncone2017, perlmutter2016, tabrez2019, lutkebohle2009, chao2010} or a human and an AI system \cite{legg2019}. Studies in this strand investigate the impact of different explainability mechanisms on various interaction scenarios and whether they are still useful when the human-robot dyad is given a concrete task. This is important, because users can focus their full attention on the explainable behaviour in the observer setting; in interaction scenarios, on the other hand, they have to divide their attention. Research in this strand is more heterogeneous, likely due to the increased design complexity of an interaction scenario. At the same time, the amount of research done, i.e., the number of papers identified, is less than the research done following the observational design above; probably because of the the above mentioned added complexity. Nevertheless, we argue that more work on this strand is needed, as we consider testing explainability mechanisms in an interaction as the gold standard for determining their utility and effectiveness.

Finally, some researchers examined participants' responses to hypothetical scenarios \cite{lee2010, poulsen2018}. The procedure in these studies is to first describe a scenario to participants in which a robot uses an explainability mechanism during an interaction with a human. Then, participants are asked to give their opinion about this interaction, which is used to determine the utility of the mechanism. This method can be very useful during the early design stages of an interaction, and can help find potential flaws in the design before spending much time implementing them on a robot. At the same time, it may be a less optimal choice for the final evaluation, especially when compared to the other methods presented above.

\section{Challenges in Explainability Research}
Shifting the focus to how results are reported in research papers on explainability, we would like to address two challenges we faced while aggregating the data for this review.

The first challenge is the large diversity and inconsistency of language used in the field. Transparency, explainability, expressivity, understandability, predictability and communicability are just a few examples of words used to describe explainability mechanisms. Authors frequently introduce their own terminology when addressing the problem of explainability. While this might allow for a very nuanced differentiation between works, it becomes challenging to properly index all the work done, not only because different authors addressing the same idea may use different terminology but also especially because different authors addressing different ideas end up using the same terminology.

Other reviews on the topic have pointed this out as well \cite{chakraborti2018b, rosenfeld2019}, and it became a challenge in our review, as we cannot ensure completeness of a keyword search based approach. The most likely cause of this is because the field is seeing rapid growth, and precise terminology is still developing.

This work tries to address this first challenge by showing how different terms are used to identify similar concepts and providing a definition that aims to be comprehensive for the surveyed papers.

The second challenge was that many authors only define the explainability mechanism they investigate implicitly. We often had to refer to the concrete experimental design to infer which mechanism was studied. While all the important information is still present in each paper, we think that explicitly stating the explainability mechanism under study can help discourse regarding explainability become much more concrete.

In extension, some authors have implemented explainability mechanisms on robotic systems that are capable of adapting their behaviour or performing some kind of learning. In many cases, these learning algorithms were unique implementations, or variations of standard algorithms, e.g., reinforcement learning, which make them very interesting. How to best incorporate an explainability mechanism into such a framework is still an open question. Unfortunately, we found that the details of the method are often underreported and that we could not extract enough data on what has been done so far. We understand that this aspect is often not the core contribution of a paper and that space is a constraint. Nevertheless, we would like to encourage future contributions to put more emphasis on how explainability mechanisms are integrated into existing learning frameworks. Technical contributions such as this could prove very valuable for defining a standardized approach to achieve explainability using embodied social agents.

\section{Open Questions}
While performing the review, we identified a set of open questions. For convenience we enumerate them here:

\begin{enumerate}
\item What are good models to predict/track human expectations/beliefs about the embodied agent’s goals and actions?
\item What are efficient learning mechanisms to include the human in the loop when building explainability into embodied agents?
\item How does the environment and embodiment influence the choice of social cues used for explainability?
\item What are good objective measures by which we can measure trust in the context of explainability?
\item Why does explainability not have a strictly positive impact on the efficiency of the interaction?
\end{enumerate}

\section{Conclusion}
Above we conducted a systematic review of the literature on explainable embodied agents. We used keyword based search to identify $32$ relevant contributions and provided a detailed analysis of them. 

First, we analysed the definitions of explainability used in each piece, highlighting the heterogeneity of existing definitions and stating our definition. In the process, we identified four main motivations that lead researchers to study explainability: (1) interactive robot/machine teaching, (2) human trust, (3) teamwork, and (4) ethical decision making. We then detailed why explainability is important for each, and identified the motivations and definition behind each of the surveyed papers. Second, we looked at social cues used as vehicles to deliver the explainability mechanism. We identified the categories of (1) speech, (2) text, (3) movement, and (4) imagery and described how each provides explainable behaviours. Third, we took stock of the algorithms used to select which part of the interaction should be made explainable. We found that only a small fraction of the work addresses this algorithmic part and most often not in sufficient detail for an in-depth analysis. We hence extended the literature in this section, to draw from other related work to provide a better overview. Fourth, we asked how the impact of explainability is measured in the identified literature. We found that most of the literature looks at three aspects: (1) trust, (2) robustness, and (3) efficiency, of which trust and robustness have received the most attention. We looked at how these aspects are measured and formulated open questions for future work.

Looking at the big picture, we identified three strands of research. The first one has partial overlap with XAI and tries to implement decision support systems on embodied agents. The second treats humans as observers and investigates what kind of explainability mechanisms can be used to make humans understand the embodied agent's inner workings. The third and final strand investigates how explainability mechanisms can be smoothly integrated into an interaction context (compared to treating humans as pure observers).

Finally, we provided a list of open questions and gaps in the literature that we identified during our analysis in the hope that further investigation will address this fascinating new domain of research.

 \section*{Acknowledgment}
This work received funding from the European Union's Horizon 2020 research and innovation programme under grant agreement No 765955 (ANIMATAS Innovative Training Network).

\bibliographystyle{ACM-Reference-Format}
\bibliography{references}

\appendix
\section{Identified Categories by Paper}
\begin{landscape}
\begin{longtable}{@{}lcccccccccccccccccl@{}}
\caption{Identified Categories by Paper}
\label{tab:core-papers}\\
\toprule
CitationKey & \multicolumn{5}{l}{Definition} &  & \multicolumn{5}{l}{Social Cues} &  & \multicolumn{4}{l}{Measurement} &  &  \multicolumn{1}{l}{Learning Paradigm} \\* 
 & \rot{transparency} & \rot{explainability} & \rot{expressivity} & \rot{other} & \rot{none} &  & \rot{Movement} & \rot{Text} & \rot{Speech} & \rot{Imagery} & \rot{other/None} &  & \rot{Trust} & \rot{Robustness} & \rot{Efficiency} & \rot{Other} &  & \rot{ML} \\
\midrule
\endfirsthead
\multicolumn{19}{c}%
{{\bfseries Table \thetable\ continued from previous page}} \\
\toprule
CitationKey & \multicolumn{5}{l}{Definition} &  & \multicolumn{5}{l}{Social Cues} &  & \multicolumn{4}{l}{Measurement} &  &  \multicolumn{1}{l}{Learning Paradigm} \\* 
 & \rot{transparency} & \rot{explainability} & \rot{expressivity} & \rot{other} & \rot{none} & & \rot{Movement} & \rot{Text} & \rot{Speech} & \rot{Imagery} & \rot{other/None} & & \rot{Trust} & \rot{Robustness} & \rot{Efficiency} & \rot{Other} & & \rot{ML}\\
\midrule
\endhead
\bottomrule
\endfoot
\endlastfoot
\citet{akash2018a} & X & . & . & . & . &  & . & X & . & . & . &  & X & . & . & . &  & . \\
\citet{akash2018b} & X & . & . & . & . &  & . & X & . & . & . &  & X & . & X & . &  & . \\
\citet{baraka2018a} & X & . & X & . & . &  & . & . & . & . & X &  & X & X & . & . &  & . \\
\citet{boyce2015} & X & . & . & . & . &  & . & . & . & X & . &  & X & . & . & . &  & . \\
\citet{brown2017} & . & . & . & . & X &  & X & . & . & X & . &  & . & . & . & X &  & . \\
\citet{chakraborti2017} & . & X & . & . & . &  & . & . & . & . & X &  & X & . & X & . &  & . \\
\citet{chao2010} & X & . & . & . & . &  & X & . & . & . & . &  & X & X & X & . &  & X \\
\citet{chen2017} & X & . & . & . & . &  & . & X & . & . & . &  & X & X & X & . &  & . \\
\citet{fischer2018} & X & . & . & . & . &  & . & . & X & . & . &  & X & . & . & . &  & . \\
\citet{floyd2016} & X & . & . & . & . &  & . & X & . & . & . &  & X & . & . & . &  & . \\
\citet{floyd2017} & X & X & . & . & . &  & . & X & . & . & . &  & X & . & . & . &  & . \\
\citet{gong2018} & . & X & . & . & . &  & . & X & . & . & . &  & X & . & . & . &  & X \\
\citet{hayes2017} & X & X & . & . & . &  & . & X & . & . & . &  & . & . & . & X &  & X \\
\citet{huang2019} & . & . & . & X & . &  & X & . & . & . & . &  & . & . & . & . &  & X \\
\citet{khoramshai2019} & . & . & . & . & . &  & . & . & . & . & X &  & . & . & . & . &  & X \\
\citet{kwon2018} & . & . & X & . & . &  & X & . & . & . & . &  & X & X & . & . &  & . \\
\citet{lamb2018} & . & . & . & . & X &  & X & . & . & . & . &  & . & X & . & . &  & . \\
\citet{lee2019} & . & . & . & X & . &  & . & . & . & . & X &  & X & . & . & . &  & . \\
\citet{legg2019} & . & . & . & . & X &  & . & . & . & X & . &  & . & X & . & . &  & X \\
\citet{lutkebohle2009} & . & . & . & . & X &  & X & . & X & . & . &  & . & . & . & X &  & X \\
\citet{perlmutter2016} & X & . & . & . & . &  & X & . & X & X & X &  & . & X & X & . &  & X \\
\citet{poulsen2018} & X & . & . & . & . &  & . & X & . & . & . &  & X & . & . & . &  & . \\
\citet{roncone2017} & X & . & . & . & . &  & . & . & X & X & . &  & . & . & X & . &  & . \\
\citet{schaefer2016} & . & . & . & X & . &  & . & . & . & X & . &  & X & . & . & . &  & . \\
\citet{sciutti2014} & . & . & . & X & . &  & X & . & . & . & . &  & . & X & . & . &  & . \\
\citet{sheikholeslami2018} & . & . & . & X & . &  & X & . & . & . & . &  & . & X & X & . &  & . \\
\citet{sreedharan2017} & . & X & . & . & . &  & . & X & . & . & . &  & . & X & . & . &  & . \\
\citet{tabrez2019} & . & X & . & . & . &  & . & . & X & . & . &  & X & . & . & . &  & X \\
\citet{wang2016a} & . & X & . & . & . &  & . & X & . & . & . &  & X & . & X & . &  & . \\
\citet{wang2016b} & . & X & . & . & . &  & . & X & . & . & . &  & . & X & X & . &  & . \\
\citet{wang2018} & . & X & . & . & . &  & . & X & . & . & . &  & X & X & . & . &  & X \\
\citet{zhou2017} & . & . & X & . & . &  & X & . & . & . & . &  & X & . & . & . &  & . \\
\citet{grigore2018} & . & . & . & X & . &  & . & . & . & . & X &  & . & X & . & . &  & . \\* \bottomrule
\end{longtable}
\end{landscape}

\section{Definitions Used by Core Papers}
\begin{longtable}{rlp{4.5cm}p{3.5cm}}
\caption{Definition of Explainability Ordered by Publication Year}
\label{tab:definition}\\
\toprule
\textbf{Year(s)} & \textbf{Author(s)} &\textbf{Definition} &\textbf{Motivation}\\
\midrule
\endfirsthead
\toprule
\textbf{Year(s)} & \textbf{Author(s)} &\textbf{Definition} &\textbf{Motivation}\\
\midrule
\endhead
\bottomrule
\endfoot
\bottomrule
\endlastfoot

2008 & \citet{fischer2008} & Robot explanations of its own actions designed to make the process and robot behaviors and capabilities accessible to the user & Trust\\\\
2009 & \citet{lutkebohle2009} & Structure verbal and non-verbal dialog to guide human actions & Machine Teaching, Predictability, Human-Robot Collaboration \\\\

2010 & \citet{chao2010} & Revealing to the teacher what is known and what is unclear & Machine Teaching, Predictability\\\\

2013 & \citet{lee2013} & Develop expectancy-setting strategies and recovery strategies to forewarn people of a robot’s limitations and reduce the negative consequence of breakdowns & Robot acceptance\\\\

2014 & \citet{sciutti2014} & Convey cues about object features (e.g., weight) to the human partner using implicit communication & Human-Robot Collaboration\\\\

2015 & \citet{boyce2015} & Display transparency information (SAT model \cite{chen2017}) in the interface of an autonomous robot & Trust\\\\

2016 & \citet{wang2016a} & Generate explanations of the robot's reasoning, communicate uncertainty and conflicting goals & Trust, Teamwork \\\\
2016 & \citet{perlmutter2016} & Communicate robot's internal processes with human-like verbal and non-verbal behaviors & Communication, Visualization, Control \\\\

2016 & \citet{schaefer2016}& 
Convey the robot's reasoning processes or intent, understanding the control allocation processes, and human engagement or reengagement strategies & Human-Robot Collaboration, Trust\\\\

2017 & \citet{floyd2016,floyd2017} & Layer that allows the agent to explain why it adapted its behaviours & Trust, Adaptation\\\\

2017 & \citet{hayes2017}, \cite{tabrez2019} & Autonomosly synthesize policy descriptions and respond to both general and target queries by human collaborators

& Human-Robot Collaboration, Control, Debug\\\\

2017 & \citet{roncone2017} & Transfer information to the human partner about its own [robot] internal state and intents
& Trust, Proficiency, Confidence, Uncertainty, Introspection\\\\

2017 & \citet{chakraborti2017}, \cite{sreedharan2017}& 
Robot’s attempt to move the human’s model to be in conformance with its own. 
& Communication, Human-Robot Collaboration, Impedance Mismatch\\\\


2017 & \citet{zhou2017} & Communicate the robot's internal state though timing & Perceived Naturalness, Human's Learning (Task Understanding)\\\\

2017 & \citet{chen2017,chen2018},\cite{akash2018a,akash2018b} 2018 & Descriptive quality of an interface pertaining to its abilities to afford an operator’s comprehension about an intelligent agent’s intent, performance, future plans, and reasoning process \cite{chen2014}.& Trust, Human's Workload \\\\ 

2018 & \citet{kwon2018} & Express robot’s incapability and communicate both what the robot is trying to accomplish and why the robot is unable to accomplish it & Robot's Acceptance, Human-Robot Collaboration\\\\

2018 & \citet{baraka2018b} & Externalize hidden information of an agent. Express robot behaviors that have a specific communicative purpose. & Human-Robot Collaboration, Control, Communication\\\\

2018 & \citet{gong2018} & Explaining robot behaviours as intention, or explicitly signalling the robot’s intentions & Teamwork, Human-Robot Collaboration\\\\

2018 & \citet{lamb2018} & Implement behavioral dynamics models based on human decision-making dynamics & Human-Robot Collaboration\\\\




2019 & \citet{legg2019} & Establish a two-way collaborative dialogue on data attributions between human and machine and express personal confidence in data attributions & Active Learning, Human-Robot Collaboration\\\\

2019 & \citet{poulsen2018} & Deciphering the behaviour of intelligent others \cite{wortham2016}, allowing ‘inspection of thoughts’.
& Ethical Decision-Making, Trust\\\\

2019 & \citet{huang2019} & Communicate information to correctly anticipate a robot’s behaviours in novel situations and building an accurate mental model of the robot's objective function & Prediction, Human-Robot Coordination\\\\

2019 & \citet{khoramshai2019} & Ensure the robot's behaviours complies with human intention, adapting generated motions (i.e., the desired velocity) to those intended by the human user & Human-Robot Collaboration\\\\
\bottomrule
\end{longtable}

\end{document}